\documentclass[letterpaper]{article} % DO NOT CHANGE THIS
\usepackage{aaai25}  % DO NOT CHANGE THIS
\usepackage{times}  % DO NOT CHANGE THIS
\usepackage{helvet}  % DO NOT CHANGE THIS
\usepackage{courier}  % DO NOT CHANGE THIS
\usepackage[hyphens]{url}  % DO NOT CHANGE THIS
\usepackage{graphicx} % DO NOT CHANGE THIS
\usepackage{natbib}  % DO NOT CHANGE THIS AND DO NOT ADD ANY OPTIONS TO IT
\usepackage{caption} % DO NOT CHANGE THIS AND DO NOT ADD ANY OPTIONS TO IT
\usepackage{algorithm}
\usepackage{algorithmic}

\usepackage{newfloat}
\usepackage{listings}
\DeclareCaptionStyle{ruled}{labelfont=normalfont,labelsep=colon,strut=off} % DO NOT CHANGE THIS
\floatstyle{ruled}
\newfloat{listing}{tb}{lst}{}
\floatname{listing}{Listing}
\usepackage[utf8]{inputenc} % allow utf-8 input
\usepackage{booktabs}       % professional-quality tables
\usepackage{amsfonts}       % blackboard math symbols
\usepackage{nicefrac}       % compact symbols for 1/2, etc.
\usepackage{microtype}      % microtypography
\usepackage{xcolor}         % colors

\usepackage{amsmath}
\usepackage{amssymb}
\usepackage{mathtools}
\usepackage{amsthm}

\usepackage[capitalize,noabbrev]{cleveref}

\theoremstyle{plain}
\newtheorem{theorem}{Theorem}[section]

\theoremstyle{definition}
\newtheorem{definition}[theorem]{Definition}

\theoremstyle{remark}
\newtheorem{remark}[theorem]{Remark}

\usepackage{tabularx}

\usepackage{svg}
\svgpath{{figs/}}

\usepackage{xspace}
\newcommand{\NAME}{{\scshape HetTree}\xspace}

\newtheorem*{example}{Example \thedefinition}

\newcommand{\xns}{$\times$} % times without space
\usepackage[russian,english]{babel}

\title{Heterogeneous Graph Neural Network on Semantic Tree}
\author {
    Mingyu Guan\textsuperscript{\rm 1},
    Jack W. Stokes\textsuperscript{\rm 2},
    Qinlong Luo\textsuperscript{\rm 2},
    Fuchen Liu\textsuperscript{\rm 2},
    Purvanshi Mehta\textsuperscript{\rm 3},
    Elnaz Nouri\textsuperscript{\rm 2},
    Taesoo Kim\textsuperscript{\rm 1}
}
\affiliations {
    \textsuperscript{\rm 1}Georgia Institute of Technology,\\
    \textsuperscript{\rm 2}Microsoft Corporation,\\  \textsuperscript{\rm 3}Lica World Inc\\
    mingyu.guan@gatech.edu, \{jstokes, qinlongluo, fuchen.liu\}@microsoft.com, purvanshi@lica.world, elnaaz@gmail.com, taesoo@gatech.edu
}

\begin{document}

\maketitle

\begin{abstract}
The recent past has seen an increasing interest in Heterogeneous Graph Neural Networks (HGNNs), since many real-world graphs are heterogeneous in nature, from citation graphs to email graphs. 
However, existing methods ignore a tree hierarchy among metapaths, naturally constituted by different node types and relation types.
In this paper, we present \NAME, a novel HGNN that models both the graph structure and heterogeneous aspects in a scalable and effective manner. 
Specifically, \NAME builds a semantic tree data structure to capture the hierarchy among metapaths.
To effectively encode the semantic tree, \NAME uses a novel subtree attention mechanism to emphasize metapaths that are more helpful in encoding parent-child relationships. 
Moreover, \NAME proposes carefully matching pre-computed features and labels correspondingly, constituting a complete metapath representation.
Our evaluation of \NAME on a variety of real-world datasets demonstrates that it outperforms all existing baselines on open benchmarks and efficiently scales to large real-world graphs with millions of nodes and edges.
\end{abstract}

\begin{links}
    \link{Code}{https://github.com/microsoft/HetTree}
    % \link{Datasets}{https://aaai.org/example/datasets}
    % \link{Extended version}{https://aaai.org/example/extended-version}
\end{links}

%%%%%% BEGIN INTRODUCTION %%%%%%
\section{Introduction}
\label{sec:introduction}
\begin{figure}[t]
    \centering
    % \includesvg[inkscapelatex=false, width=\linewidth]{email_graph}
    \includegraphics[width=\linewidth]{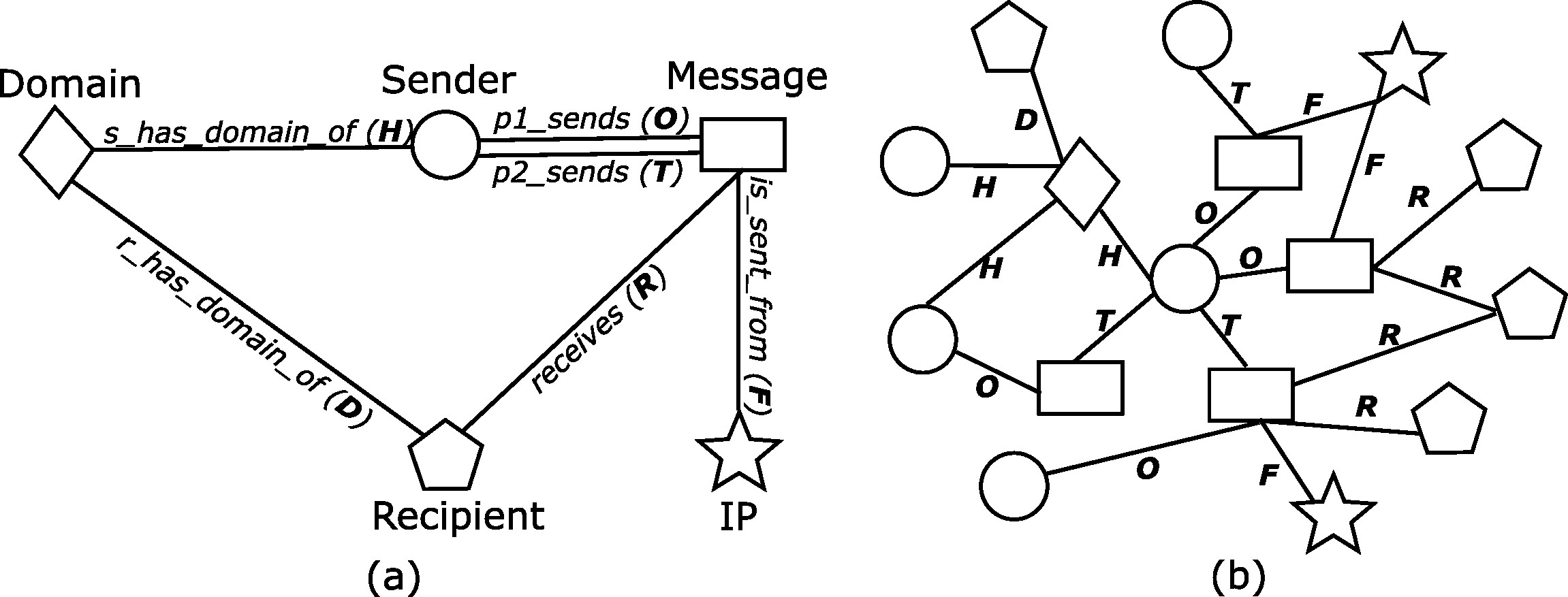}
    \caption{(a) Relational scheme of a heterogeneous email graph (b) An example of the email graph.}
    \label{fig:email_graph}
\end{figure}

Graph neural networks (GNNs) have been widely explored in a variety of domains from social networks to molecular properties~\cite{PinnerSage2020, Park2019, sun2022does}, where graphs are usually modeled as homogeneous graphs. 
However, real-world graphs are often heterogeneous in nature~\cite{OGB, HGB}. 
For example, as shown in \cref{fig:email_graph}(a), a heterogeneous email graph can include multiple types of nodes - Domain, Sender, Recipient, Message, and IP Address - and the relations among them. 
Moreover, multiple relations can exist between two entities in complex heterogeneous graphs. 
For example, a \emph{Sender} node can be a P1 sender and/or P2 sender of a \emph{Message}, 
where the P1 sender denotes the entity that actually sent the message while an email application displays the P2 sender as the ``From'' address. 
For example, if Bob sends an email on behalf of Alice, the email appears to originate from Alice, making Alice the P2 sender and Bob the P1 sender.
As a result, there are two relations between Sender and Message in \cref{fig:email_graph}(a), \textit{p1\_sends} and \textit{p2\_sends}.

To better understand real-world heterogeneous graphs, various heterogeneous graph neural networks (HGNNs) have been proposed. 
The most well-known approaches are the metapath-based methods~\cite{RGCN, HAN, MAGNN}, which first aggregate neighbor representations along each metapath at the node level and then aggregate these representations across metapaths at the metapath (semantic) level. 
However, metapath-based approaches often involve manual effort to select a subset of meaningful metapaths, because the node-level aggregation along each metapath is computationally expensive~\cite{MAGNN, HAN}.
Other models that do not apply the metapath method, such as HetGNN~\cite{HetGNN} and HGT~\cite{HGT}, carefully encode representations for different node types and/or relation types in heterogeneous graphs. 
These fine-grained embedding methods often utilize multi-layer message-passing techniques as in traditional GNNs, thus facing scalability issues. 
To efficiently model real-world web-scale graphs, researchers and practitioners have explored various ways to scale HGNNs.
Sampling-based methods sample sub-graphs with different strategies~\cite{HGT,HetGNN}, while others use model simplification to execute feature propagation as a pre-processing stage before training~\cite{NARS, SeHGNN}.

However, existing HGNNs fail to account for a \textit{tree hierarchy among the metapaths}. 
A metapath represents an ordered sequence of composite relationships connecting distinct or identical node types (\cref{def:metapath}). 
For instance, as illustrated in \cref{fig:email_graph}(a), the metapath $Sender \xrightarrow{p1\_sends} Message \xrightarrow{is\_sent\_from} IP$ is intuitively more closely associated with $Sender \xrightarrow{p1\_sends} Message$ than with $Sender \xrightarrow{s\_has\_domain\_of} Domain$ due to a greater overlap in node types and relationships. 
This overlap can be conceptualized as a parent-child relationship, where the parent metapath serves as a prefix to its child metapaths. 
Consequently, these parent-child relationships naturally form a tree hierarchy among the metapaths.

The exploration of tree structures in heterogeneous graphs is not a new topic, as several tree-based HGNNs have investigated this concept. However, existing tree-based HGNNs~\cite{SHGNN, HetGTCN, T-GNN} primarily utilize tree structures to capture the \textit{local topology of nodes}. These approaches subsequently employ attention mechanisms for semantic aggregation to integrate information across metapaths. Despite these efforts, they do not account for the parent-child relationships or hierarchical organization inherent among metapaths (\cref{sec:tree_hgnns}).

Moreover, to augment the data usage, label utilization has been widely adopted in GNNs~\cite{SLE, UniMP, GAMLP, LEGNN}.
These methods leverage ground truth labels from the training set and propagate them through the graph structure as inputs to the model. 
However, existing approaches either completely separate feature learning from label learning, combining them only in the final stage, or treat feature and label vectors equivalently by projecting them to the same latent space.
While such strategies may be effective for homogeneous graphs, they are less suitable for heterogeneous graphs, where features and labels can propagate along distinct metapaths, and \textit{features and labels through the same metapath are more related}.

In this paper, we present \NAME, a scalable HGNN that extracts a unified tree representation on metapaths, i.e., \textit{semantic tree}, from the input graph and proposes a novel tree aggregation with subtree attention to encode the resulting semantic tree, in which propagated labels are carefully matched with corresponding features based on metapaths. 

To scale efficiently on web-scale graphs, \NAME follows the model simplification approach that simplifies heterogeneous feature aggregation as a pre-processing stage.
Meanwhile, label aggregation of the target node types for each metapath is also executed. 
\NAME then builds a semantic tree to capture the hierarchy among metapaths.
Instead of separating feature learning from label learning~\cite{SLE, GAMLP} or treating features and labels equally by projecting them to the same latent space~\cite{SeHGNN}, \NAME proposes to match them carefully on corresponding metapaths, which provides more accurate and richer information between node features and labels.

To encode the resulting semantic tree, \NAME uses a novel \textit{subtree attention} mechanism to emphasize children nodes that are more helpful in encoding parent-child relationships.
Existing tree encoding techniques~\cite{SHGNN, HetGTCN, T-GNN} aggregate children nodes by weighting the contribution of children nodes based on similarity to the parent node.
However, in the semantic tree, this tree encoding fails to capture the entire parent-child hierarchy by only considering the parent node.
Hence, subtree attention in \NAME models a broader parent-child structure while enhancing
correlations, bringing a better representation for each metapath.

In summary, we make the following contributions.
\begin{itemize}
\item We observe that existing HGNNs ignore the hierarchy of the metapath features. \NAME takes a radically new approach to encoding metapath hierarchy by building a \textit{semantic tree} for both pre-computed features and labels. 
\item \NAME proposes a novel tree aggregation with 
\textit{subtree attention} to encode the semantic tree 
structure.
For better label usage, \NAME matches pre-computed features and labels correspondingly,
which constitutes a complete representation of a metapath.
\item We conduct extensive experiments on five open graph datasets 
as well as a real-world commercial email dataset. 
The results demonstrate that \NAME can outperform the state-of-the-art architectures on all datasets with low computation and memory overhead.
\end{itemize}
%%%%%% END OF INTRODUCTION %%%%%%

%%%%%% BEGIN RELATED WORK %%%%%%
\section{Related Work}\label{sec:related}
%
% \subsection{Graph Neural Networks}
\noindent\textbf{Graph Neural Networks.}
\label{sec:gnns}
Graph Neural Networks (GNNs) are neural networks that take input structured as graphs. 
The fundamental task in a GNN is to generate the representation of graph entities, such as nodes 
and edges, in a $d$-dimensional space, referred to as the embedding of the entity. 
To generate the 
embedding, GNNs usually use a multi-layer feature 
propagation followed by a neural network to combine 
structural information from the graph structure and 
the input features. Various GNN architectures exist
today, but they differ in how the information is 
aggregated and transformed~\cite{Hamilton2017, 
GCN2016, GAT2018, GraphSage2017}. Nevertheless, a main 
problem with these classic GNNs is that they are hard 
to scale due to the feature propagation performed at 
each layer of the neural networks. Hence, many sampling 
methods~\cite{GraphSage2017, FastGCN2018, LADIES2019, 
zou2019layer} have been proposed to reduce both 
computation and memory complexity by using only a subset of nodes and edges. Besides sampling, other 
methods~\cite{SGC, SIGN, S2GC} simplify models by 
making feature propagation an offline
stage, so that this computation-intensive process only 
needs to be executed once and is not involved during the 
training. 

\noindent\textbf{Heterogeneous Graph Neural Networks.}
\label{sec:hgnns}
To extend GNN from homogeneous graphs to heterogeneous graphs, 
various heterogeneous GNN architectures (HGNNs) have been explored.
The most general approach in HGNNs is the so-called 
metapath-based method~\cite{MAGNN, HAN}, where the 
feature propagation is performed based on semantic 
patterns and an attention mechanism is usually applied at 
both the node level for each metapath and the semantic level 
across metapaths. Other models~\cite{RGCN, HetGNN, HGT, HGB} 
encode graph heterogeneity at a more fine-grained 
level using the multi-layer, message-passing framework common in GNNs, 
where different weights are learned for distinct entity types. 

However, HGNNs also inherit the scalability problem from traditional homogeneous GNNs. Hence, 
sampling~\cite{HGT} and model simplification~\cite{NARS} 
have also been explored in the heterogeneous graph learning 
domain. NARS~\cite{NARS} first applies the scaling approach proposed by SIGN~\cite{SIGN} on heterogeneous graphs, 
which samples multiple relational subgraphs using different sets of relations and then treats them as homogeneous graphs. 
SeHGNN~\cite{SeHGNN} computes averaged metapath features separately and applies Transformer-like attention to learn metapath features. 
However, simply applying the Transformer ignores the hierarchy among metapath features and thus results in sub-optimal results.

\noindent\textbf{Tree-based HGNNs.}
\label{sec:tree_hgnns}
Recent work~\cite{HGTM23} utilizes a topic tree as a regularizer, i.e., log-likelihood terms, for text decoding on document graphs.
For general heterogeneous graphs, a few HGNNs have explored tree structure based on the local topology of nodes.
T-GNN~\cite{T-GNN} and SHGNN~\cite{SHGNN} construct hierarchical tree structures at the node level, where each tree represents a metapath and each level 
of the tree contains nodes of a certain type. 
Similarly, HetGTCN~\cite{HetGTCN} also constructs a tree hierarchy for each node, where tree nodes at the $k^{th}$ level are \textit{k-hop} neighbors of the root node.
% How they encode tree-structured data
To encode tree-structured data, i.e. parent-child relationships among tree nodes, they either use a weighted sum aggregator~\cite{T-GNN, HetGTCN} or compute weights using an attention mechanism for each parent-child pair~\cite{SHGNN, HetGTCN}. 
These methods utilize the tree structure to aggregate neighbor information for each metapath first and then use semantic attention~\cite{HAN} to aggregate metapath representations to obtain the final node representation.
However, \textit{the tree hierarchy among metapaths} has not been explored.

\noindent\textbf{Label Utilization.}
\label{sec:label_utilization}
Label utilization has been commonly applied in graph representation 
learning. 
In general, partially observed labels in the training set are propagated 
through the network structure to generate label representations, combined with the feature representations to generate the final representations of graph entities~\cite{LabelPropagation, LabelUsage, SLE}.
To avoid label leakage and overfitting, UniMP~\cite{UniMP} randomly masks the training nodes for 
each epoch. GAMLP~\cite{GAMLP} modifies label propagation 
with residual connections to each hop to alleviate the label leakage issue.

However, these methods either completely separate feature learning 
and label learning and only combine them at the end, or they simply add 
the features and label vectors together as propagation information, 
which may result in good performance in homogeneous graphs but not 
in the case of heterogeneous graphs, since the features and labels 
are related by the corresponding metapaths.
%%%%%% END OF RELATED WORK %%%%%%

%%%%%% BEGIN PRELIMINARY %%%%%%
\section{Preliminary}\label{sec:preliminary}
In this section, we provide formal definitions of important 
terminologies related to \NAME. 

\begin{definition} \textbf{Heterogeneous Graph.}
A heterogeneous graph is denoted as $\mathcal{G} = 
(\mathcal{V}, \mathcal{E}, \mathcal{O}, \mathcal{R})$,
where each node $v \in \mathcal{V}$ and edge $e \in 
\mathcal{E}$ are associated with a node mapping function 
$\tau(v): \mathcal{V} \rightarrow \mathcal{O}$ from 
node set $\mathcal{V}$ to node type set $\mathcal{O}$, and 
a edge mapping function $\phi(e): \mathcal{E} \rightarrow 
\mathcal{R}$ from edge set $\mathcal{E}$ to relation set 
$\mathcal{R}$, respectively.
\end{definition}

\begin{example}
\cref{fig:email_graph}(a) shows the relational scheme of 
a heterogeneous email graph while \cref{fig:email_graph}(b) 
shows an illustrative example. 
It is composed of five types of nodes: $Domain$,$Sender$, $Message$, 
$Recipient$, $IP$, and six types of relations: 
\textit{s\_has\_domain\_of(\textbf{H})}, \textit{r\_has\_domain\_of(\textbf{D})}, 
\textit{p1\_sends(\textbf{O})}, 
\textit{p2\_sends
(\textbf{T})}, \textit{receives(\textbf{R})},
\textit{is\_sent\_from(\textbf{F})}.
\end{example}

\begin{figure*}[t]
    \centering
    % \includesvg[inkscapelatex=false, width=0.9\textwidth]{htree_construction}
    \includegraphics[width=0.9\linewidth]{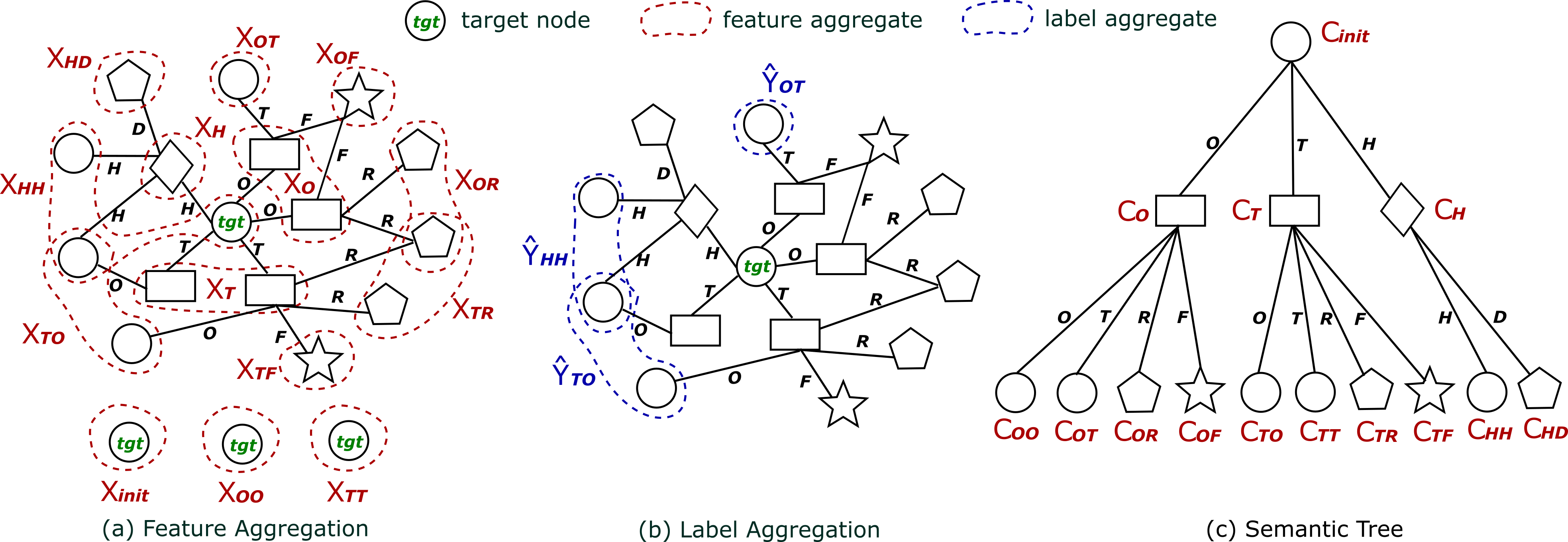}
    \caption{(a) The offline process of feature aggregation. The center node is the target $Sender$ node and features are aggregated for all metapaths $\mathcal{P}^k$ up to hop $k$, where $k=2$ in this example. 
    (b) The offline process of label aggregation on partially observed labels in the training set. 
    (c) Semantic tree with height of $k$ for \textit{Sender} nodes in the email graph. A tree node $C_P$ represents metapath $P$, where $P \in \mathcal{P}^k$.} 
    \label{fig:semantic_tree}
\end{figure*}

\begin{definition}\label{def:metapath}
\textbf{Metapath.}
A metapath $P$ is a path that describes a composite relation 
$R = R_1 \circ R_2 \circ \cdots \circ R_l$ between node types 
$O_1$ and $O_{l+1}$, where $\circ$ denotes the composition operator 
on relations. $P$ is denoted as $O_1 \xrightarrow{\text{$R_1$}} 
O_2 \xrightarrow{\text{$R_2$}} \cdots \xrightarrow{\text{$R_l$}} 
O_{l+1}$, which is abbreviated as $R_1R_2 \cdots R_l$. A special
metapath where no relation is present but includes only a node 
type $O$ is simply denoted as $O^{init}$ (abbreviated as $init$ 
when $O$ is specified). The set of metapaths ending with node
type $O$ excluding $P=O^{init}$ is denoted as $\mathcal{P}_O$.
The set of metapaths up to hop $k$ is denoted as $\mathcal{P}^k$,
where $l \leq k, \forall P=R_1R_2 \cdots R_l \in \mathcal{P}^k$. 
% A metapath instance $p$ of metapath $P$ is a sequence of nodes 
% and edges that following schema of $P$. 
Metapath-based neighbors $\mathcal{N}^v_P$ of node $v$ along 
metapath $P$ are the set of nodes that are connected with 
node $v$ via metapath $P$. Note that when $\tau(v) = O$ and 
$P$ ends with $O$, $\mathcal{N}^v_P$ can include $v$ itself.
\end{definition}

\begin{remark}
Most of metapath-based HGNNs denote a metapath $P = O_1 \xrightarrow{\text{$R_1$}} 
O_2 \xrightarrow{\text{$R_2$}} \cdots \xrightarrow{\text{$R_l$}} 
O_{l+1}$ as $O_1O_2 
\cdots O_{l+1}$ for short. We note that this notation fails
to differentiate metapaths that have different relation compositions
but the same node types along the metapaths, as multiple relations 
can be present between two node types in a heterogeneous graph.
\end{remark}

\begin{example}
In \cref{fig:email_graph}, a sender can be connected to 
a recipient through three 2-hop metapaths: $Sender \xrightarrow{O} 
Message \xrightarrow{R} Recipient$ ($OR$), $Sender \xrightarrow{T} 
Message \xrightarrow{R} Recipient$ ($TR$), $Sender \xrightarrow{H} 
Domain \xrightarrow{D} Recipient$ ($HD$). Moreover, let $S = \{OR, 
TR, HD\}$, then we have $S \subset \mathcal{P}^k$, for any $k \geq 2$.
\end{example}

\begin{definition}\textbf{Semantic Tree.}
A semantic tree $T_O$ with depth of $k$, for node 
type $O$, is composed of tree nodes $\mathcal{C} = \{C_P, \forall P \in 
\mathcal{P}^k\}$ and relation edges $\mathcal{R}$. The root node 
represents the metapath $P = O^{init}$, denoted as $C_{O^{init}}$ 
(abbreviated as $C_{init}$ when $O$ is specified). A non-root node 
$C_{R_1R_2 \cdots R_l}$ represents the metapath from the 
root node $C_{init}$ to them via relation edges $R_1R_2 \cdots R_l$.
The root node $C_{init}$ is the parent of all nodes $C_{R_1}$ 
at depth $1$ of the semantic tree $T_O$.
A node $C_{R_1R_2 \cdots R_l}$ with depth $\geq 2$ has a parent 
node $C_{R_1R_2 \cdots R_{l-1}}$, and they are connected by edge $R_l$. 
\end{definition}

\begin{example}
A semantic tree with depth of $2$ for \textit{Sender} nodes in 
the heterogeneous email graph is shown in \cref{fig:semantic_tree}, where the root node $C_{init}$
represents metapath \textit{init} and other nodes represent 
the metapath from the root node to them. For example, node 
$C_{OF}$ is connected with the root node $C_{init}$ by relation 
edges $O$ and $F$ in order.
\end{example}

%%%%%% END OF PRELIMINARY %%%%%%

%%%%%% BEGIN METHODOLOGY %%%%%%
\section{Methodology}\label{sec:methodology}
In this section, we describe a novel heterogeneous tree graph neural 
network, \NAME, for scalable and effective heterogeneous graph learning.
\NAME consists of three major components: offline feature aggregation and semantic tree
construction (\cref{sec:feat_lb_agg}), metapath 
feature transformation (\cref{sec:meta_feat_trans}), and semantic 
tree aggregation (\cref{sec:tree_agg}). 

\subsection{Offline Aggregation and Semantic Tree Construction}
\label{sec:feat_lb_agg}
As a pre-processing stage, node features and labels are aggregated 
to prepare for training. In node classification tasks, initial node features are normally associated with every node 
(if not, we can use external graph embedding algorithms like 
ComplEx~\cite{ComplEx} to generate them),
while labels are often only associated with nodes of target node types that 
need to be classified. For example, in the email dataset we collect, the task is to classify \textit{Sender} nodes as a compromised email account or not, 
and only \textit{Sender} nodes are associated with labels. Moreover, as 
mentioned in \cref{sec:label_utilization}, only labels of training 
nodes can be used as part of the input features, and the problem of label 
leakage in label utilization should be well addressed. 
We also construct a novel semantic tree structure to organize the aggregation results and capture the hierarchy of metapaths, which can be leveraged in semantic tree aggregation (\cref{sec:tree_agg}).

\subsubsection{Feature Aggregation}
\label{sec:feat_agg}
\cref{fig:semantic_tree}(a) shows the process of offline feature 
aggregation. Unlike existing metapath-based methods~\cite{HAN, MAGNN}, 
where feature aggregation is involved with model learning such as projection 
and attention, \NAME performs feature aggregation as a pre-processing step.
Existing methods manually select metapaths with domain knowledge. The choice 
saves computation complexity but also results in information loss. As the 
feature aggregation happens offline and involves no parameter learning, 
which is much less expensive than existing approaches, we use all metapaths 
up to hop $k$, where $k$ is a user-defined parameter. 
For example, when $k = 2$ as shown in \cref{fig:semantic_tree}(a), 
feature aggregation is conducted on 14 metapaths
($init$, $O$, $T$, $H$, $OO$, $OT$, $OR$, $OF$, $TT$, $TO$, $TR$, $TF$, $HH$, $HD$). Possible aggregators
include but are not limited to $mean$, $sum$, $max$, and $min$, and 
we use the $mean$ aggregator in this paper  for both feature and label aggregation. For each node $v$,
we compute a set of aggregated features $\mathcal{X}^v$:
\begin{equation}
\mathcal{X}^v = \{ X^v_P = agg(\{x^u, \forall u \in \mathcal{N}^v_P\}), 
\forall P \in \mathcal{P}^k \} 
\end{equation}
where $X_P^v$ represents the aggregated feature for node $v$ along metapath $P$, and
$agg$ is the aggregation function ($mean$ by default).

\subsubsection{Label Aggregation}
% \noindent\textit{Label Aggregation}
\label{sec:lb_agg}
\cref{fig:semantic_tree}(b) shows the process of offline label 
aggregation. The label aggregation process is very similar to the feature 
aggregation process described in \cref{sec:feat_agg}, except for two 
differences: first, since only labels from nodes with target node type 
$O_{tgt}$ (node type to be classified) in the training set can be used, 
the label aggregation is only conducted for 
metapaths $\mathcal{P}_{O_{tgt}}^k$ where they end at node type $O_{tgt}$; 
second, feature aggregation applies to all nodes in $\mathcal{N}^v_P$ 
including $v$ itself, while $v$ is excluded during label 
aggregation to avoid label leakage. For example, when $k = 2$ as shown in \cref{fig:semantic_tree}(b), label aggregation is conducted on 3 metapaths
($OT,TO, HH$) excluding the center target node, respectively. Note that only 
labels from nodes in the training set are used for label aggregation, and 
zero vectors are used for nodes in the non-training set. Specifically, 
for each node $v$, we compute a set of aggregated features $\hat{\mathcal{Y}}^v$:
\begin{equation}
\small
\begin{split}
\hat{y^v} & =  \left\{ \begin{array}{lll}
    y^v, & \mbox{if  } v \in training\_set \\ 
    \textbf{0}, & \mbox{otherwise}
    \end{array}\right. \\
\hat{\mathcal{Y}}^v & = \{ \hat{Y}^v_P = agg(\{\hat{y}^u, \forall u \in 
\mathcal{N}_v^P \setminus \{v\}\}), \forall P \in \mathcal{P}_{O_{tgt}}^k \}
\end{split}
\end{equation}

where $y^v$ is the ground truth label for node $v$, $\hat{y}^v$ is the label 
used in label aggregation for node $v$, and $\hat{Y}_P^v$ is the aggregated label 
along metapath $P$ for node $v$.

\subsubsection{Semantic Tree Construction}
% \noindent\textit{Semantic Tree Construction.}
\label{sec:tree_construction}
We can construct a semantic tree $T_O$ for node type $O$ with tree nodes 
$\{C_P, \forall P \in \mathcal{P}^k\}$. $C_{init}$ is the root node and 
a non-root node $C_{R_1R_2 \cdots R_l}$ represents the metapath from the 
root node $C_{init}$ to itself via relation edges $R_1R_2 \cdots R_l$.
The parent of all 1-hop tree nodes $C_{R_1}$ is $C_{init}$, which are 
at depth 1 of $T_O$. Starting from depth 2, $C_{R_1R_2 \cdots R_l}$ is 
connected with its parent node $C_{R_1R_2 \cdots R_{l-1}}$ via edge $R_l$.
By constructing the semantic tree, the hierarchy among metapaths can be 
captured, which provides the model structural information of the 
metapaths. Moreover, the semantic tree is also used as the 
underlying data structure for semantic tree aggregation discussed in 
\cref{sec:tree_agg}, where the metapath features are aggregated following the tree structure in a bottom-up way. \cref{fig:semantic_tree}(c) shows an illustration of the semantic 
tree for the \textit{Sender} node type.

\subsection{Metapath Feature Transformation}
% \noindent\textbf{Metapath Feature Transformation.}
\label{sec:meta_feat_trans}
After obtaining aggregated features and labels for metapaths, we transform 
them to the same latent space. This is due to the aggregated features for 
metapaths being generated from raw features of metapath-based neighbors 
with different node types, which may have different initial spaces.
Instead of separating the transformation of features and labels as in 
existing methods~\cite{SLE, GAMLP}, \NAME automatically matches and 
concatenates ($\parallel$) the aggregated features and labels of the same metapath 
$P$ for $P \in \mathcal{P}_{O_{tgt}}$. This gives the model more accurate label information of its metapath-based neighbors by designating the aggregated 
labels with corresponding metapaths. Specifically, for all $P \in \mathcal{P}^k$, we compute the metapath 
features $\mathcal{M}$ as
\begin{equation}
\small
    \mathcal{M} = \{ M_P = \left\{ \begin{array}{lcl}
    MLP(X_P \parallel \hat{Y}_P), & \mbox{if  } P \in \mathcal{P}_{O_{tgt}}^k \\ 
    MLP(X_P), & \mbox{otherwise}
\end{array}\right.\}.
\end{equation}

\subsection{Semantic Tree Aggregation}
% \noindent\textbf{Semantic Tree Aggregation.}
\label{sec:tree_agg}
As discussed in \cref{sec:tree_construction}, we construct a 
semantic tree $T$, and each tree node $C_P$ represents a metapath $P$, 
where $P \in \mathcal{P}^k$. Since we also obtain metapath features 
$\mathcal{M}$, we can associate each tree node $C_P$ 
with $M_P$ correspondingly. Note that the semantic tree structure is the 
same for all nodes with the same node type in a heterogeneous graph, so 
the target nodes (to be classified) can easily be batched. We now have 
tree-structured metapath features, and the hierarchical relationship 
between metapath features needs to be well modeled when aggregating them.

The tree aggregation in \NAME is conducted in a bottom-up fashion. As it gets closer and closer to the target node as the process proceeds, the semantic tree aggregation can gradually emphasize those metapaths that contribute more to the local subtree structure, i.e., the parent-child relationship.
To calculate the encoded representation $Z_P$ for each metapath node in the semantic tree, 
\NAME applies a novel \textit{subtree attention} mechanism to
aggregate the children nodes thus encoding the local subtree structure. 
Unlike existing tree encoding methods~\cite{TreeLSTM, T-GNN, SHGNN, HetGTCN} that use either a simple weighted-sum aggregator or attention mechanism to emphasize parent tree nodes, \NAME proposes the subtree attention to encode both the parent and children representation and uses it to emphasize the hierarchical correlation between metapaths. 
Specifically, let $\mathcal{P}^{child}_P$ be the set of metapaths that $\{C_Q, Q \in \mathcal{P}^{child}_P\}$ are the set of children 
nodes of $C_P$ in the semantic tree, \NAME computes a \textit{subtree reference} as $S_P = MLP(M_P \parallel \sum_{Q \in \mathcal{P}^{child}_P} M_Q)$, for each metapath $P$ in the semantic tree.
Then, the weight coefficient $a_Q$ of each children node $C_Q$ can be calculated as:
\begin{equation}
    \alpha_Q = \frac{exp(\delta(W_P \cdot [S_P \parallel Z_Q]))}
                {\sum_{B \in \mathcal{P}^{child}_P} exp(\delta(W_P \cdot [S_P 
                \parallel Z_B]))}.
\end{equation}
where $\delta$ is the activation function, $W_P$ is a learnable projection vector 
for metapath $P$ and $\parallel$ stands for concatenation. Then, we can finally 
compute the encoded representation $Z_P$ for parent node $C_P$ by aggregating encoded representation of children nodes as:
\begin{equation}
    Z_P = M_P + \delta(\sum_{Q \in \mathcal{P}^{child}_P} \alpha_Q \cdot 
                Z_Q).
\end{equation}

\begin{figure}[t]
    \centering
    % \includesvg[inkscapelatex=false, width=\linewidth]{subtree_attention}
    \includegraphics[width=\linewidth]{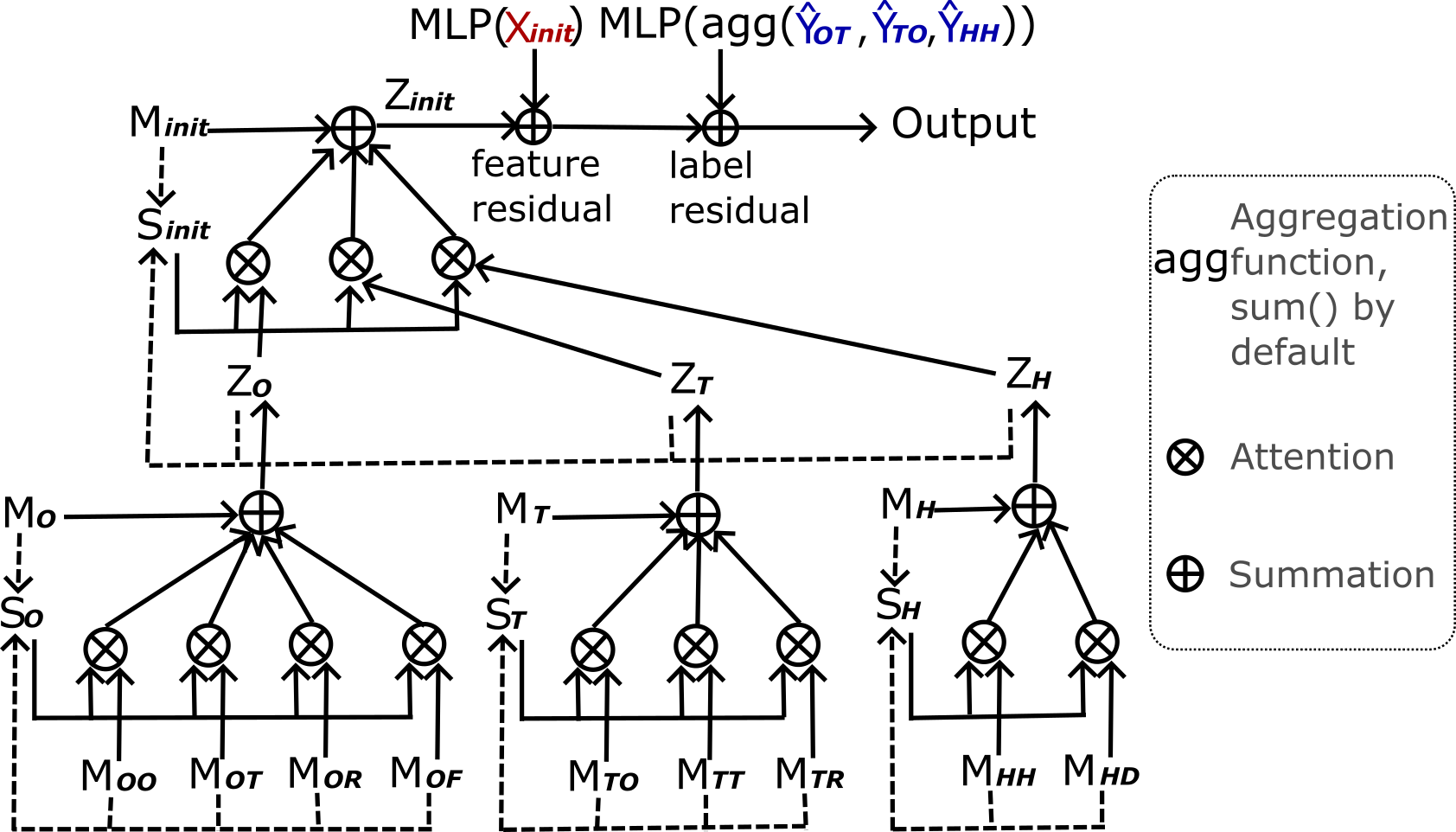}
    \caption{Semantic tree aggregation in \NAME. }
    \label{fig:het_tree}
\end{figure}

After the semantic tree aggregation has finished from bottom to top, the sum of 
metapath representations will be used as the final representation of the semantic 
tree. Moreover, we add a feature residual and a label residual to further 
emphasize the initial features and labels aggregated from the metapath-based 
neighbors. Specifically, 
\begin{equation}
\small
\begin{split}
    Y_{pred} \quad = & \quad MLP(\sum_{P \in \mathcal{P}^k} Z_P) + MLP(X_{init}) \\
         & \quad + MLP(agg.(\{\hat{Y}_P, \forall P \in \mathcal{P}_{O_{tgt}}^k \})).
\end{split}
\end{equation}

\noindent where $agg.$ is an aggregation function, which can be $mean$, $sum$, $max$, 
$min$, etc. An illustration of the semantic tree aggregation process is shown in 
\cref{fig:het_tree}, following the same example in \cref{fig:semantic_tree}.

%%%%%% END OF METHODOLOGY %%%%%%

%%%%%% BEGIN EXPERIMENTS %%%%%%
\section{Experiments}\label{sec:experiments}
We conduct extensive experiments on six heterogeneous graphs to answer the following questions.

\begin{description}
  \item[Q1.] How does \NAME compare to the state-of-the-art overall on open benchmarks?
  \item[Q2.] How does \NAME perform in a practical compromised account detection on a noisy email graph?
  \item[Q3.] How does each component of \NAME contribute to the performance gain?
  \item[Q4.] Is \NAME practical w.r.t. running time and memory usage? 
\end{description}

\begin{table*}[t]
\centering
\small
% \begin{tabular}{ccccccccc}
\begin{tabularx}{\textwidth}{XXXXXXXXX}
    \hline
     & \multicolumn{2}{c}{DBLP} & \multicolumn{2}{c}{IMDB} & \multicolumn{2}{c}{ACM} & \multicolumn{2}{c}{Freebase} \\
    \hline
     & Macro-F1 & Micro-F1 & Macro-F1 & Micro-F1 & Macro-F1 & Micro-F1 & Macro-F1 & Micro-F1  \\
    \hline
    RGCN & 91.52$\pm$0.50 &  92.07$\pm$0.50 & 58.85$\pm$0.26 & 62.05$\pm$0.15 & 91.55$\pm$0.74 & 91.41$\pm$0.75 & 46.78$\pm$0.77 & 58.33$\pm$1.57 \\
    HAN & 91.67$\pm$0.49 & 92.05$\pm$0.62 & 57.74$\pm$0.96 & 64.63$\pm$0.58 & 90.89$\pm$0.43 & 90.79$\pm$0.43 & 21.31$\pm$1.68 & 54.77$\pm$1.40 \\
    HetGNN & 91.76$\pm$0.43 & 92.33$\pm$0.41 & 48.25$\pm$0.67 & 51.16$\pm$0.65 & 85.91$\pm$0.25 & 86.05$\pm$0.25 & - & - \\
    MAGNN & 93.28$\pm$0.51 & 93.76$\pm$0.45 & 56.49$\pm$3.20 & 64.67$\pm$1.67 & 90.88$\pm$0.64 & 90.77$\pm$0.65 & - & - \\
    HGT & 93.01$\pm$0.23 & 93.49$\pm$0.25 & 63.00$\pm$1.19 & 67.20$\pm$0.57 & 91.12$\pm$0.76 & 91.00$\pm$0.76 & 29.28$\pm$2.52 & 60.51$\pm$1.16 \\
    HGB & 94.01$\pm$0.24 & 94.46$\pm$0.22 & 63.53$\pm$1.36 & 67.36$\pm$0.57 & 93.42$\pm$0.44 & 93.35$\pm$0.45 & 47.72$\pm$1.48 & 66.29$\pm$0.45 \\
    SeHGNN & 95.06$\pm$0.17 & 95.42$\pm$0.17 & 67.11$\pm$0.25 & 69.17$\pm$0.43 & 94.05$\pm$0.35 & 93.98$\pm$0.36 & 51.87$\pm$0.86 & 65.08$\pm$0.66 \\
    \hline
    \textbf{\NAME} & \textbf{95.34$\pm$0.17} & \textbf{95.64$\pm$0.15} & \textbf{68.43$\pm$0.31} & \textbf{70.92$\pm$0.29} & \textbf{94.26$\pm$0.20} & \textbf{94.19$\pm$0.20} & \textbf{52.35$\pm$0.96} & \textbf{66.39$\pm$0.40}\\
    \hline
\end{tabularx}
\caption{Experimental Results of \NAME and baselines over four graphs in the HGB benchmark. "-" means that the model runs out of memory on the corresponding graph.}
\label{tbl:hgb-res}
\end{table*}

\textbf{Experimental Setup.} All of the experiments were 
conducted on a machine with dual 12-core Intel Xeon Gold 
6226 CPU, 384 GB of RAM, and one NVIDIA Tesla A100 80GB GPU.
The server runs 64-bit Red Hat Enterprise Linux 7.6 with CUDA 
library v11.8, PyTorch v1.12.0, and DGL v0.9.

\textbf{Datasets.}
We evaluate \NAME on four graphs from the HGB~\cite{HGB} benchmark: DBLP, IMDB, ACM, and Freebase, a citation graph \textit{Ogbn-Mag} from the OGB benchmark~\cite{OGB} and a real-world email dataset collected from a commercial email platform. We summarize the six graphs in ~\cref{tbl:datasets}.

\begin{table}[t]
\centering
\small
\begin{tabularx}{\linewidth}{XXXXXX}
 \toprule
 \textbf{Graph} & \textbf{Nodes} & \textbf{Edges} & \begin{tabular}{@{}c@{}} \textbf{Node} \\ \textbf{Types}\end{tabular} & \begin{tabular}{@{}c@{}} \textbf{Relation} \\ \textbf{Types}\end{tabular} & \textbf{Classes} \\
 \midrule
  DBLP     &  26.1K   & 239.5K    & 4 & 6  & 4\\
  IMDB     &  21.4K   & 86.6K     & 4 & 6  & 5\\
  ACM      &  10.9K   & 547.8K & 4 & 8  & 3\\
  Freebase &  180.0K  & 1.0M   & 8 & 36 & 7\\
  Mag &  1.9M    & 21.1M & 4 & 5  & 349\\
  Email    &  7.8M    & 34.9M          & 5 & 6  & 2 \\
 \bottomrule
\end{tabularx}
\caption{Statistics of datasets.}
\label{tbl:datasets}
\end{table}

\textbf{Baselines.}
For the four graphs from the HGB benchmark, we compare the \NAME results to the results reported in the HGB paper~\cite{HGB} as well as a state-of-the-art work SeHGNN~\cite{SeHGNN}. For \textit{Ogbn-Mag}, we compare the \NAME with top-performing methods from either the baseline paper or the leaderboard of OGB~\cite{OGB}. 
We use the unified metrics chosen by the benchmarks for a fair comparison, i.e., the HGB benchmark uses F1 scores while the OGB benchmark uses accuracy as metrics.
For the email dataset, we compare \NAME with the best-performing baseline SeHGNN.
All experimental results reported are averaged over five random seeds.

\textbf{Ethics and Broader Impacts.} 
This work was reviewed and approved by independent experts in Ethics, Privacy, and Security. For the email dataset, all users' identities were anonymized twice, and the map from the second anonymized user IDs to the first anonymized user IDs was deleted. Furthermore, the data was handled according to GDPR regulations. 

\subsection{Experiments on Open Benchmarks}\label{sec:benchmarks}
To answer \textbf{Q1}, we compare the performance of the proposed \NAME model to state-of-the-art models on five heterogeneous graphs from two open benchmarks - HGB~\cite{HGB} and OGB~\cite{OGB}.\\
\noindent\textbf{Performance on HGB Benchmark.}
\cref{tbl:hgb-res} shows results that compare \NAME with the best-performing baselines on four datasets from the HGB benchmark.
\NAME outperforms the baselines on all graphs in terms of both Macro-F1 and Micro-F1 scores.
For datasets in the HGB benchmark, which share similar medium-scale sizes and have uniformly preprocessed input node features, we observe that \NAME derives greater benefits from its semantic tree aggregation mechanism on more complex tasks involving a larger number of classes.
As shown in \cref{tbl:hgb-res}, \NAME has more performance gain on IMDB and Freebase with 5 and 7 classes, respectively, compared with DBLP and ACM with 3 and 4 classes, respectively.
This can be attributed to \NAME's semantic tree aggregation that learns more information, i.e., the hierarchy among metapaths, which is ignored by the other baselines.

\noindent\textbf{Performance on Ogbn-Mag Dataset.}
We also evaluate \NAME on a large-scale citation graph, Ogbn-Mag~\cite{OGB}, with millions of nodes in \cref{tab:ogbn-mag-res}.
We report results \textit{without} self-enhanced techniques like multi-stage training~\cite{li2018deeper,sun2020multi, LEGNN}, which are orthogonal to \NAME and can be incorporated for additional benefits.
The results show that \NAME maintains its benefits on large graphs and outperforms all baselines. 

\begin{table}[h]
\centering
\small
\begin{tabular}{c|cc}
    \hline
    Methods & Validation Accuracy & Test Accuracy \\
    \hline
    RGCN & 48.35 $\pm$ 0.36& 47.37 $\pm$ 0.48 \\
    HGT & 51.24 $\pm$ 0.46 & 49.82 $\pm$ 0.13 \\
    NARS & 53.72 $\pm$ 0.09 & 52.40 $\pm$ 0.16 \\
    LEGNN & 54.43 $\pm$ 0.09 & 52.76 ± 0.14 \\
    GAMLP  & 55.48 $\pm$ 0.08 & 53.96 $\pm$ 0.18 \\
    SeHGNN & 56.56 $\pm$ 0.07 & 54.78 $\pm$ 0.17 \\
    \hline
    \textbf{\NAME}  & \textbf{57.31 $\pm$ 0.15} & \textbf{55.54 $\pm$ 0.17}  \\
    \hline
\end{tabular}
\caption{Detection accuracy of the \NAME model and other baselines for the Ogbn-Mag dataset.}
\label{tab:ogbn-mag-res}
\end{table}

\subsection{Experiments on Commercial Email Graph}
Besides the open benchmark datasets, we also collect a large email dataset from a commercial email platform to answer \textbf{Q2}, for lack of public alternatives. 
In this experiment, a subsample of real-world email data is used, which contains five types of entities - senders, recipients, domains, IP addresses, and messages.
The task is to predict if the sender is legitimate or compromised, given its domain, messages, recipients of the message, recipients' domains, and message IP addresses.
Compromised accounts may send various types of malicious emails, such as phishing emails, malware attachments, and spam.
The email dataset has highly imbalanced classes where legitimate email accounts are much more than compromised accounts, as in the real-world scenario.

\begin{figure}[h!]
    \centering
    \includegraphics[width=0.7\linewidth]{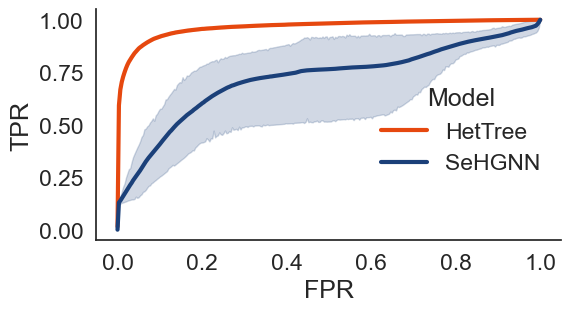}
    \caption{ROC Curves for the email dataset. The error bars for \NAME are tiny.}
    \label{fig:email_roc}
\end{figure}

Since the email dataset contains binary labels, we can construct a Receiver Operating Characteristic (ROC) curve for the models.
The ROC curve for the detection of compromised emails is presented in ~\cref{fig:email_roc} for the email dataset, and the accuracies are shown in ~\cref{tab:email}.
These results demonstrate that while both the best-performing baseline SeHGNN~\cite{SeHGNN} and \NAME can achieve high accuracy in classification, HetTree can distinguish better between positive and negative classes.

\begin{table}[h]
\centering
\small
\begin{tabular}{c|c|c}
    \hline
    Methods & Val Accuracy & Test Accuracy \\
    \hline 
    SeHGNN & 97.22 $\pm $0.02 &  97.26 
    $\pm$ 0.02  \\
    \hline 
    \textbf{\NAME} & \textbf{98.48$\pm$0.01} & \textbf{98.48 $\pm$ 0.02}  \\
    \hline 
\end{tabular}
\caption{Detection accuracy for the email dataset.}
\label{tab:email}
\end{table}

\subsection{Ablation Study}\label{sec:ablation}
We next evaluate whether adding the subtree attention component and using labels from the training set really help or not to answer \textbf{Q3}. The test F1 scores or accuracy of \NAME are evaluated on IMDB, ACM, and Ogbn-Mag compared with its three variants: \textit{"weighted-sum}", "\textit{parent-att}" and "\textit{no-label}". Variant \textit{weighted-sum} removes the proposed subtree attention but uses a weighted child-sum like TreeLSTM~\cite{TreeLSTM}. Variant \textit{parent-att} removes the proposed subtree attention but computes weights of children nodes using attention on the parent node, which is the tree-encoding method used in SHGNN~\cite{SHGNN}. Variant \textit{no-label} does not use labels as extra inputs. 
We note that substituting the entire semantic tree aggregation with conventional metapath aggregation based on attention mechanism is what the baseline SeHGNN~\cite{SeHGNN} does. Since we compare \NAME with SeHGNN in all benchmarks, we do not include it in the ablation study.

The results in \cref{tbl:ablation} show that each component is effective for \NAME. We notice that datasets exhibit distinct sensitivities to individual components. 
The subtree attention results in more performance gain on IMDB, which could be attributed to the sparsity of the graph.
It demonstrates that subtree attention can capture the metapath hierarchy, compared to other tree encoding mechanisms.
% The GRU unit results in more performance gain on Ogbn-Mag than on IMDB and ACM, which could be attributed to the more complicated semantic tree structure. 
Ogbn-Mag is more sensitive to label utilization, which could be attributed to the large number of classes of labels that provide richer information through propagation.

\begin{table}[h]
\centering
\small
\begin{tabular}{cccc}
 \toprule
   & \textbf{IMDB} & \textbf{ACM} & \textbf{Ogbn-Mag} \\
 \midrule
   & Micro-F1 & Micro-F1 & Accuracy \\
 \midrule
  \NAME     & \textbf{70.92$\pm$0.29} & \textbf{94.19$\pm$0.20}  & \textbf{55.54$\pm$0.17} \\
  \textit{weighted-sum} & 69.70$\pm$0.24   & 93.54$\pm$0.23  & 55.38$\pm$0.21\\
  \textit{parent-att}   & 69.69$\pm$0.41   & 93.83$\pm$0.18  & 54.87$\pm$0.26\\
  \textit{no-label} & 70.11$\pm$0.25  & 93.41$\pm$0.23 & 52.93$\pm$0.12\\
 \bottomrule
\end{tabular}
\caption{Effectiveness of each component of \NAME.}
\label{tbl:ablation}
\end{table}

\subsection{Computation Cost Comparison}\label{sec:computation-cost}
We next investigate the computational cost of \NAME in terms of epoch time and memory footprint to answer \textbf{Q4}.
We select three performant models - HAN~\cite{HAN}, HGB~\cite{HGB}, and SeHGNN~\cite{SeHGNN} - to compare with \NAME on four graphs in the HGB benchmark.
For fair comparison, we use a 2-layer structure for HAN and HGN, and 2-hop feature propagation for SeHGNN and \NAME.
The result in \cref{fig:time-memory} shows that \NAME incurs the lowest computational cost in terms of both running time and memory usage across three datasets from the HGB benchmark.

\begin{figure}[h]
  \centering
  \input{figs/time-memory}
  \caption{
    Epoch time and memory usage on HGB datasets.
  }
  \label{fig:time-memory}
\end{figure}

%%%%%% END OF EXPERIMENTS %%%%%%

%%%%%% BEGIN CONCLUSION %%%%%%

\section{Conclusion}\label{sec:conclusion}
In this paper, we present a novel HGNN, \NAME, based on the observation that existing HGNNs 
ignore a tree hierarchy among metapaths, which is naturally constituted by 
different node types and relation types. 
\NAME builds a semantic tree structure to capture the hierarchy among metapaths 
and proposes a novel subtree attention mechanism to encode the semantic tree.
Compared with existing tree-encoding techniques that weight the contribution
of children nodes based on similarity to the parent node, subtree attention in \NAME can model the broader local structure of parent nodes and children nodes.
The evaluation shows that \NAME can outperform state-of-the-art baselines on open benchmarks and efficiently scale to large real-world graphs.

%A future direction is to generalize the semantic tree structure to not only scalable HGNNs but also HGNNs with multi-layer aggregation, which performs a more fine-grained encoding on semantics and the semantic tree structure could bring additional benefits. 

%%%%%% END OF CONCLUSION %%%%%%

\bibliography{references}

\end{document}